\documentclass[sigconf, 9pt]{acmart}
\usepackage{booktabs}

\usepackage{amsfonts}
\usepackage{amsmath}
\usepackage{colortbl}
\usepackage{graphicx,subcaption,float}
\usepackage{enumerate}
\usepackage{caption}
\usepackage{algorithm}
\usepackage[noend]{algpseudocode}
\usepackage{hyperref,url,xspace}
\usepackage{arydshln}
\DeclareMathOperator*{\argmin}{\arg\!\min}

\algdef{SE}[DOWHILE]{Do}{doWhile}{\algorithmicdo}[1]{\algorithmicwhile\ #1}
\usepackage{algpseudocode}
\usepackage{algorithm}
\usepackage{etoolbox}

\makeatletter
\newcommand{\algrule}[1][.2pt]{\par\vskip.5\baselineskip\hrule height #1\par\vskip.5\baselineskip}

\usepackage[title]{appendix}
\usepackage{multirow}

\begin{document}
\title{Mobile Sensor Data Anonymization }

\author{Mohammad~Malekzadeh}
\affiliation{%
  \institution{Queen Mary University of London,~UK}
}
\email{m.malekzadeh@qmul.ac.uk}

\author{Richard G. Clegg}
\affiliation{%
  \institution{Queen Mary University of London,~UK}
}
\email{r.clegg@qmul.ac.uk}

\author{Andrea Cavallaro}
\affiliation{%
  \institution{Queen Mary University of London,~UK}
}
\email{a.cavallaro@qmul.ac.uk}

\author{Hamed  Haddadi}
\affiliation{%
  \institution{Imperial College London,~UK}
}
\email{h.haddadi@imperial.ac.uk}

\begin{abstract}
Motion sensors such as accelerometers and gyroscopes measure the instant acceleration and rotation of a device, in three dimensions. Raw data streams from motion sensors  embedded in portable and wearable devices may reveal private information about users without their awareness.
For example, motion data might disclose the weight or gender of a user, or enable their re-identification.
To address this problem, we propose an on-device transformation of sensor data to be shared for specific applications, such as monitoring selected daily activities, without revealing information that enables user identification. We formulate the anonymization problem using an information-theoretic approach and propose a new multi-objective loss function for training deep autoencoders. 
This loss function helps minimizing user-identity information as well as data distortion to preserve the application-specific utility. The training process regulates the encoder to disregard user-identifiable patterns and tunes the decoder to shape the output independently of users in the training set. The trained autoencoder can be deployed on a mobile or wearable device to anonymize sensor data even for users who are not included in the training dataset. Data from 24 users transformed by the proposed anonymizing autoencoder lead to a promising trade-off between utility and privacy, with an accuracy for activity recognition above 92\% and an accuracy for user identification  below 7\%.

\end{abstract}
\copyrightyear{2019} 
\acmYear{2019} 
\setcopyright{acmlicensed}
\acmConference[IoTDI '19]{International Conference on Internet-of-Things Design and Implementation}{April 15--18, 2019}{Montreal, QC, Canada}
\acmBooktitle{International Conference on Internet-of-Things Design and Implementation (IoTDI '19), April 15--18, 2019, Montreal, QC, Canada}
\acmPrice{15.00}
\acmDOI{10.1145/3302505.3310068}
\acmISBN{978-1-4503-6283-2/19/04}
\begingroup
\mathchardef\UrlBreakPenalty=10000
\maketitle
\endgroup
\section{Introduction}
\begin{figure}[t]
\centerline{\includegraphics[scale=.52]{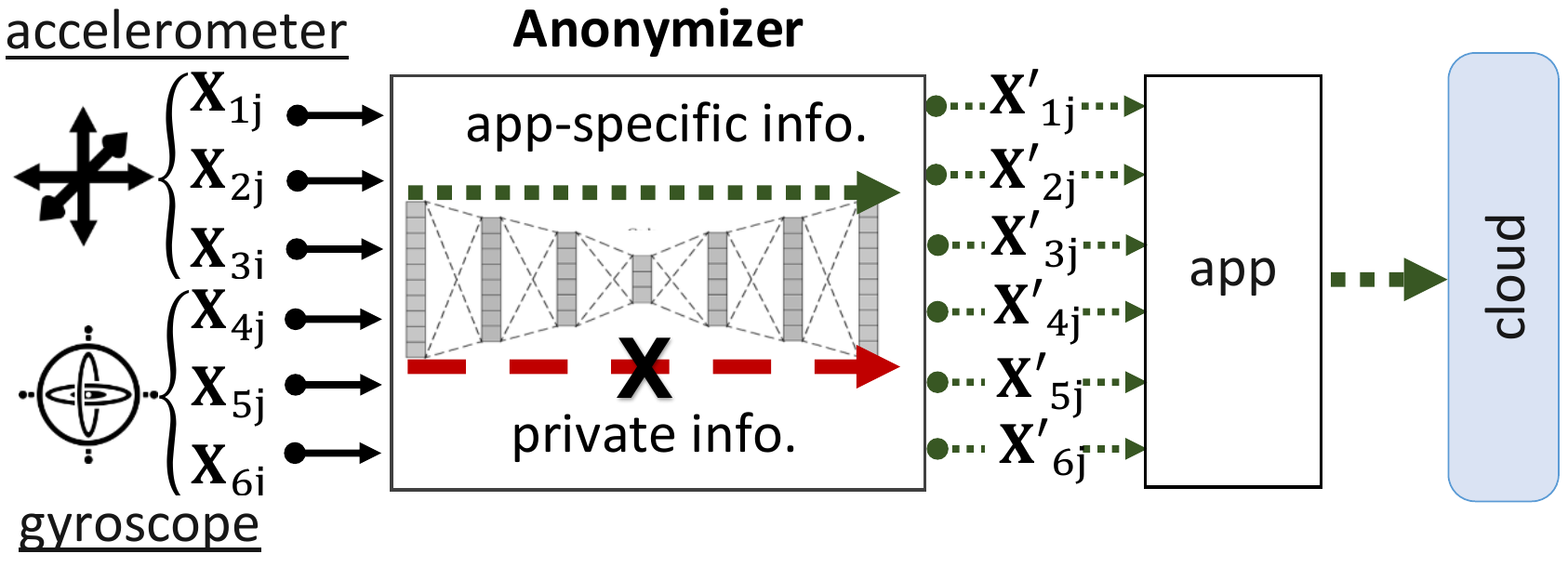}}
 \caption{The Anonymizer is a pre-trained autoencoder that transforms raw data before they are shared with an (untrusted) app to enable a service-specific inference that does not reveal private information about the user. KEY - $\mathbf{X}_{sj}$: raw data generated by sensor ${s}$ at time ${j}$; $\mathbf{X'}_{sj}$: corresponding anonymized data  after transformation.}
    \label{fig:big_pic}
\end{figure} 

Motion data from the sensors in mobile and wearable devices can reveal private information about users without their awareness. For instance, motion patterns can be used to create fine-grained behavioral profiles of users that reveal their identity~\cite{neverova2016learning}. We are interested in designing an on-device privacy-preserving approach to share with apps transformed sensor data in order to prevent the exposure of sensitive information unrelated to the service while simultaneously preserving the service-specific utility~(see Figure~\ref{fig:big_pic}).

Approaches for privacy-preserving data release include differentially private mechanisms~\cite{dwork2010differential} and information theoretic frameworks~\cite{sankar2013utility}.  {\em Differential privacy}~\cite{wang2015review} offers a privacy guarantee for access to private datasets, but it is not applicable to continuously released sensor data. In fact, a private mechanism for publishing sensitive data needs to aggregate all users' data~\cite{dp2017learning} and, in our scenario, we do not trust  data aggregators. Moreover, we want to run the mechanism on user devices, but the local version of differential privacy~\cite{duchi2013local, liu2017deeprotect} is unsuitable in this case. Time series such as sensor data present recurring patterns in consecutive temporal windows and,  unless considerable noise is added to each window that would eliminate the utility of the data, applying the same differentially private mechanism to all windows does not provide a privacy guarantee~\cite{tang2017privacy}. Instead, frameworks based  on {\em information theory}~\cite{ma2015information} consider  as the measure of privacy the mutual information between the released data and the latent information that can be inferred from data. Under this framework we do not necessarily need to design a noise addition mechanism and we can  remove or at least reduce  private information while keeping useful service-specific information~\cite{rassouli2018optimal}.

To design a data release mechanism that simultaneously satisfies utility and privacy constraints, we use adversarial approaches to train deep autoencoders~\cite{makhzani2015adversarial}. Using adversarial training~\cite{edwards2015censoring,tripathy2017privacy}, we approximate the mutual information by estimating the posterior distribution of private variables, given the released data. Moreover, we anonymize data locally and  define a mechanism that can be shared across users, whereas existing solutions need a trusted party to access user personal data to offer a reliable distortion mechanism~\cite{liu2017deeprotect, osia2018deep, xiao2018information} or need users to participate in a privacy-preserving training mechanism~\cite{abadi2016deep}. 

We formulate  the sensor data anonymization problem as an optimization process based on information theory and propose a new way of training deep autoencoders. Inspired by recent advances in adversarial training to discover from raw data useful representations for a specific task~\cite{makhzani2015adversarial}, we propose a new multi-objective loss function to train deep autoencoders~\cite{masci2011stacked}. The loss function regulates the transformed data to keep as little information as possible about user identity, subject to a minimal distortion  to preserve utility, which in our case is that of an activity recognition service.

Unlike other approaches~\cite{malekzadeh2018protecting, raval2019olympus, sankar2013utility, hamm2017minimax, huang2017context, tripathy2017privacy}, our training process not only regulates the encoder to consider exclusively task-specific features in the data, but also shapes the final output independently of the specific users in the training set. This process leads to a generalized model that can be applied to new data of unseen users, without user-specific re-training. We evaluate the efficiency and utility-privacy trade-off of the proposed mechanism and compare it with other methods on an activity recognition dataset\footnote{Code and data are available at:
\url{https://github.com/mmalekzadeh/motion-sense}}.


\section{Related Work}
\label{sec:related}

Adversarial learning enables us to approximate, using generative adversarial networks (GANs)~\cite{goodfellow2014generative}, the underlying distribution of data or to model, using variational autoencoders (VAE)~\cite{kingma2013auto}, data with  well-known distributions. These techniques can be applied to quantify mutual information for optimization problems~\cite{huang2017context,   osia2018deep,  tripathy2017privacy, edwards2015censoring} and can be used to remove sensitive information from latent low-dimensional representations of the data, e.g.~removing text from images~\cite{edwards2015censoring}. An optimal privacy mechanism can be formulated as a game between two players, a privatizer and an adversary, with an iterative minimax algorithm~\cite{huang2017context}. 
Moreover, the service provider can share a feature extractor based on an initial training set that is then re-trained by the user on their data and then sent back to the service provider~\cite{shamsabadi2018distributed, osia2018deep}.

In our work, we do not assume the existence of a trusted data aggregator to perform anonymization for end users. We assume we only have access to a public dataset for training a general anonymization model. The trained anonymizer should  generalize to new unseen users, because it is impractical for all users to provide their data for the training.

The feature maps of a convolutional autoencoder have the ability to extract patterns and dependencies among data points and have shown good performance in time series analysis~\cite{yang2015deep}. Autoencoders compress the input into a low-dimensional latent representation and then reconstruct the input from this representation. Autoencoders  are usually trained by minimizing the differences  (e.g.~mean squared error or cross entropy) between the input and its reconstruction~\cite{masci2011stacked}. 
The bottleneck of the autoencoder forces the training process to capture the most descriptive patterns in the data (i.e. the main factors of variation of the data) in order to generalize the model and prevent undesirable memorization~\cite{gehring2013extracting,bengio2009learning}. An effective way to train an autoencoder is to randomly corrupt~\cite{vincent2008extracting} or replace~\cite{malekzadeh2018replacement} the original input and force the model to refine it in the reconstruction. In this way, a well-trained autoencoder captures prominent and desired patterns in the data and ignores noise or undesired patterns~\cite{vincent2008extracting}. Moreover, a latent representation can be learned that removes some  meaningful patterns from the data to reduce the risk of inferring sensitive information~\cite{malekzadeh2018replacement}.

Only considering the latent representation produced by the encoder and leaving intact the decoder with information extracted from the training data  offer only limited protection~\cite{edwards2015censoring, liu2017deeprotect}. Considering the decoder's output leads to a more reliable data protection~\cite{malekzadeh2018protecting, raval2019olympus}. In this paper, we consider outputs from both the encoder and decoder of an autoencoder  for data transformation. We also consider a distance function as an adjustable constraint on the transformed data to control the amount of data distortion and help tune the privacy-utility trade-off for different applications.

\section{Sensor Data Anonymization}\label{sec:anon}

We aim to produce a data transformation mechanism to anonymize mobile sensor data so that the user specific motion patterns, that are highly informative about user's identity, cannot be captured by an untrusted app that has access to the sensor to recognize a set of $B$ required activities. Thus, we consider users' identity, that can be inferred from user specific motion patterns, as their sensitive data.
We use the concept of mutual information to quantify how much can be inferred about a particular variable from a data set.  We wish to minimize the amount the data changes but remove the ability to infer private information from the data.\footnote{As notation we use capital bold-face, e.g. $\mathbf{X}$, for random variables (univariate or multivariate) and lowercase bold-face, e.g. $\mathbf{x}$, for  an instantiation; roman typestyle, e.g. $\mathrm{I}$, for operations or functions; lowercase math font, e.g. $i$, for indexing; and capital math font, e.g. $M$, for specific numbers such as the size of a vector.}

\subsection{Anonymization function}

Let sensor component ${s}$ (e.g. the $z$ axis value of the gyroscope sensor) at sampling instant ${j}$,  generate  $\mathbf{X}_{{sj}}\in \mathbb{R}$. Let the time series generated by $M$ sensor components in a time-window of length $W$,  be represented by matrix $\mathbf{X}\in\mathbb{R}^{M\times W}$, with
$\mathbf{X}=( \mathbf{X}_{{sj}}) $.
Let $N$ be the number of users and $\mathbf{U} \in \{0,1\}^{N}$ be a variable representing the identity of the user; a one-hot vector of length $N$, a vector with $1$ in the ${k}$-th place and $0$ in all other places if user ${k}$ generated the data being considered.  Let the current activity that generates $\mathbf{X}$ be $\mathbf{T}\in \{0,1\}^{B}$; a one-hot vector of length $B$ with the one in position $b$ if the current activity is the $b$-th activity. Finally, we define the  data with the user's identifiable information obscured as the \textit{anonymized sensor data}, $\mathbf{X'}$.

\begin{figure}[t!]
	\centering
	\includegraphics[scale=0.26 ]{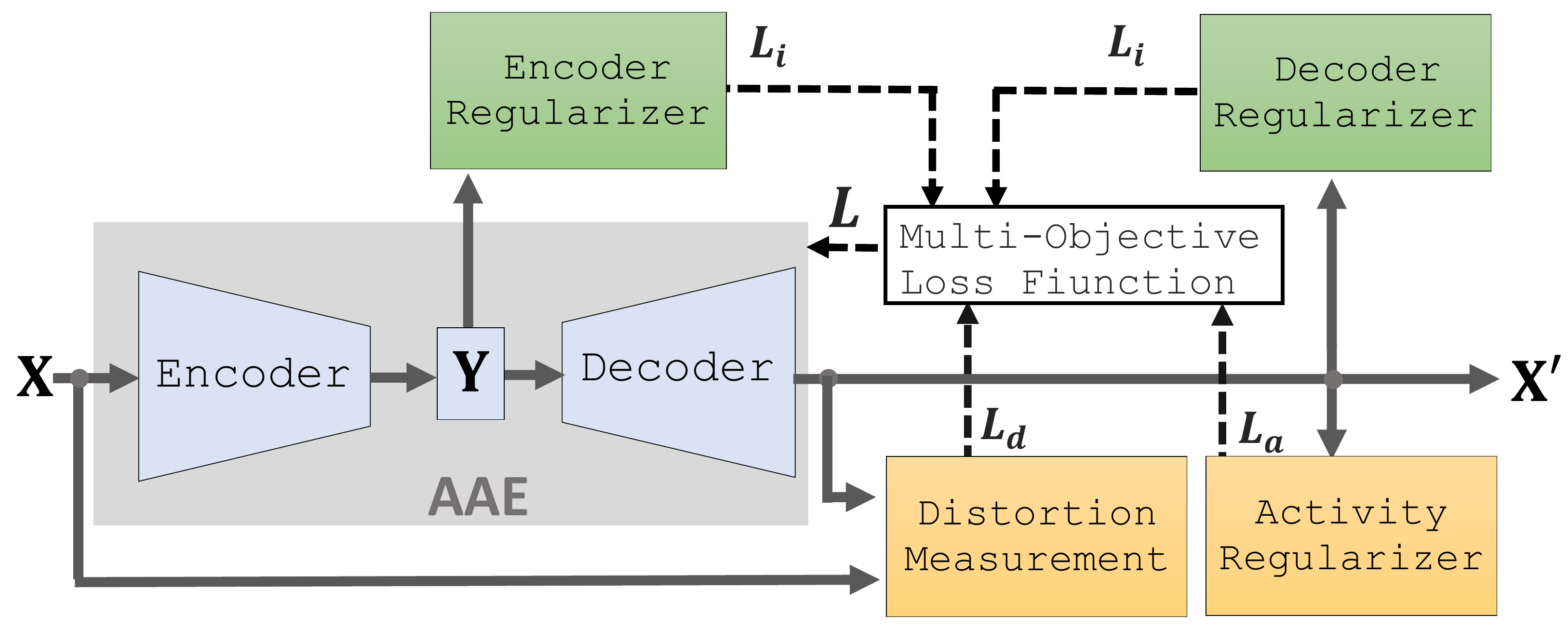}
	\caption{\label{fig:framework} The  losses involved in the training procedure. After training, the Anonymizing AutoEncoder~(AAE), or Anonymizer, runs on the device as interface between sensor data and (untrusted) apps. KEY -- Solid lines: data flow; dashed lines: loss functions. $\mathbf{X}$: raw input data; $\mathbf{Y}$: low-dimensional representation of the input data; $\mathbf{X'}$: transformed data; $\mathrm{L}_{i}$: identity loss; $\mathrm{L}_{a}$: activity loss; $\mathrm{L}_{d}$: distortion loss function; $\mathrm{L}$: overall loss function for training the AAE.}
\end{figure}

Let $\mathrm{I}(\cdot;\cdot)$ be the mutual information function,  $\mathrm{d}(\cdot,\cdot)$ a distance function between two time series\footnote{In the specific implementation of this paper, we choose as  $\mathrm{d}(\cdot,\cdot)$ the mean squared error, $\mathrm{MSE}$, between raw data and the corresponding transformed data. One can choose any other distance functions based on the tasks at hand.}, $\mathrm{A}(.)$ a data transformation function and  $\mathbf{X}$ the data we want to anonymize. We define the fitness function $\mathrm{F}(.)$  as 
\begin{equation}\label{eq:anon}
    \mathrm{F}\left(\mathrm{A}\left(\mathbf{X}\right)\right)
    =
    \beta_{i}
    \mathrm{I}\left(\mathbf{U};\mathrm{A}\left(\mathbf{X}\right)\right)
    -
    \beta_{a}\mathrm{I}\left(\mathbf{T}; \mathrm{A}\left(\mathbf{X}\right)\right)
    +
    \beta_{d} \mathrm{d}\left(\mathbf{X},\mathrm{A}\left(\mathbf{X}\right)\right),
\end{equation}
where the non-negative weight parameters $\beta_i$, $\beta_a$ and $\beta_d$ determine the trade-off between privacy and utility.

Let us define the {anonymization function},  $\mathcal{A}(\cdot)$, as 
\begin{equation}
    \mathcal{A}\left(\mathbf{X}\right) = \argmin_{\mathrm{A}\left(\mathbf{X}\right)}{ \mathrm{F}\left(\mathrm{A}\left(\mathbf{X}\right)\right)}.
\end{equation}
%
%
such that the optimal $\mathcal{A}(\cdot)$ transforms $\mathbf{X}$ into  $\mathbf{X'}=\mathcal{A}(\mathbf{X})$, which  contain as little information as possible associated to the  identity of the user (minimum $\mathrm{I}(\mathbf{U};\mathbf{X'})$), while maintaining sufficient information to discriminate the activity (maximum $\mathrm{I}(\mathbf{T};\mathbf{X'})$) and minimizing the distortion of the original data (minimum $\mathrm{d}(\mathbf{X},\mathbf{X'})$). 

As we cannot practically search over all possible anonymization functions,  we consider a deep neural network and look for the optimal parameter set through training. To approximate the required mutual information terms, we reformulate the optimization problem in (\ref{eq:anon}) as a neural network optimization problem and train an anonymizing autoencoder~(AAE) based on adversarial training.

\subsection{Architecture}

%
Let $\mathrm{A}(\mathbf{X}; \theta)$ be an autoencoder neural network, where  $\theta$ is the parameter set and  $\mathbf{X}$ is the input vector to be transformed into the output vector $\mathbf{X}'$ with the same dimensions.
The network {optimizer} finds the optimal parameter set $\theta^{*}$ by searching the space of all the possible parameter sets, $\Theta$, as:
\begin{equation}\label{eq:id_act}
\theta^{*}=\argmin_{ \theta \in \Theta} \beta_i\mathrm{I}\left(\mathbf{U};\mathrm{A}\left(\mathbf{X}; \theta\right)\right)
-
\beta_a\mathrm{I}\left(\mathbf{T}; \mathrm{A}\left(\mathbf{X}; \theta\right)\right)
+
\beta_d\mathrm{d}\left(\mathbf{X}, \mathrm{A}\left(\mathbf{X}; \theta\right)\right)
\end{equation}
where, $\mathcal{A}(\cdot; \theta^{*})$ is the optimal estimator for a general $\mathcal{A}(\cdot)$ in (\ref{eq:anon}).

%
We obtain $\theta^{*}$  using backpropagation with stochastic gradient descent and a  multi-objective loss function. We also determine values of $\beta_i$, $\beta_a$ and $\beta_d$ as the trade-off between utility and privacy through cross validation over the training dataset.

Figure~\ref{fig:framework} shows the framework for the training of the AAE. 
The {Encoder} maps $\mathbf{X}$ into an identity concealing low-dimensional latent representation $\mathbf{Y}$ by getting feedback from a pre-trained classifier, the Encoder Regularizer, which penalizes the Encoder if it captures information corresponding to  $\mathbf{U}$ into $\mathbf{Y}$. The {Decoder} outputs a reconstruction of the input, $\mathbf{X'}$, from the $\mathbf{Y}$, and gets feedback from other pre-trained classifiers, the Decoder Regularizer and the Activity Regularizer, respectively.  
\begin{figure}[t!]
	\includegraphics[scale=0.36]{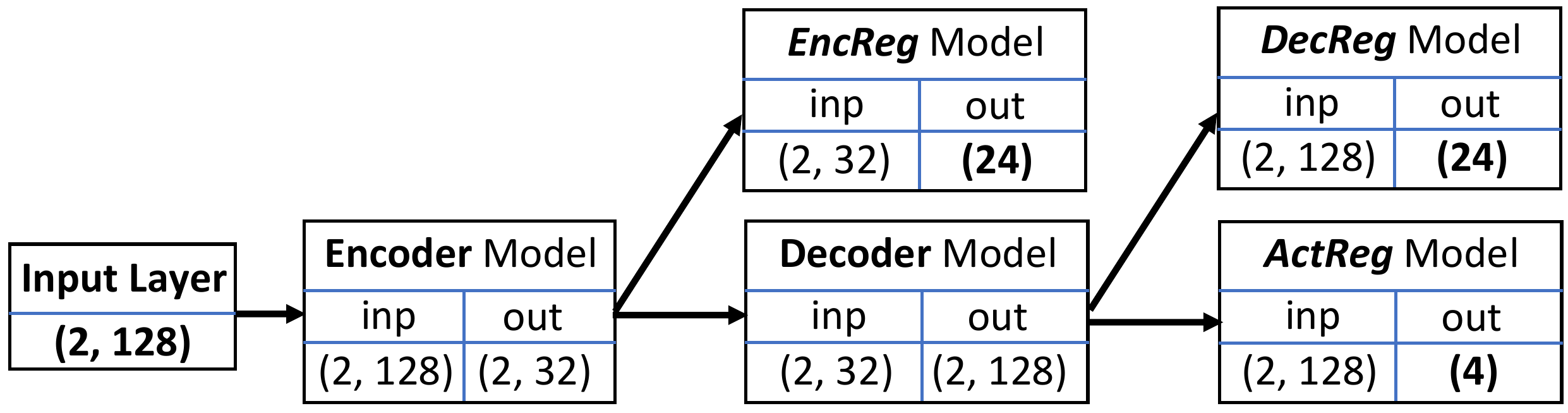}
	\caption{Implementation of the models shown in Figure~\ref{fig:framework} for a  dataset with 24 users and 4 activities. KEY --  \textit{EncReg}: Encoder Regularizer; \textit{DecReg}:  Decoder Regularizer; \textit{ActReg}: Activity Regularizer. }
	\label{fig:apm1}
\end{figure}

\begin{figure}[t!]
	\includegraphics[scale=0.5]{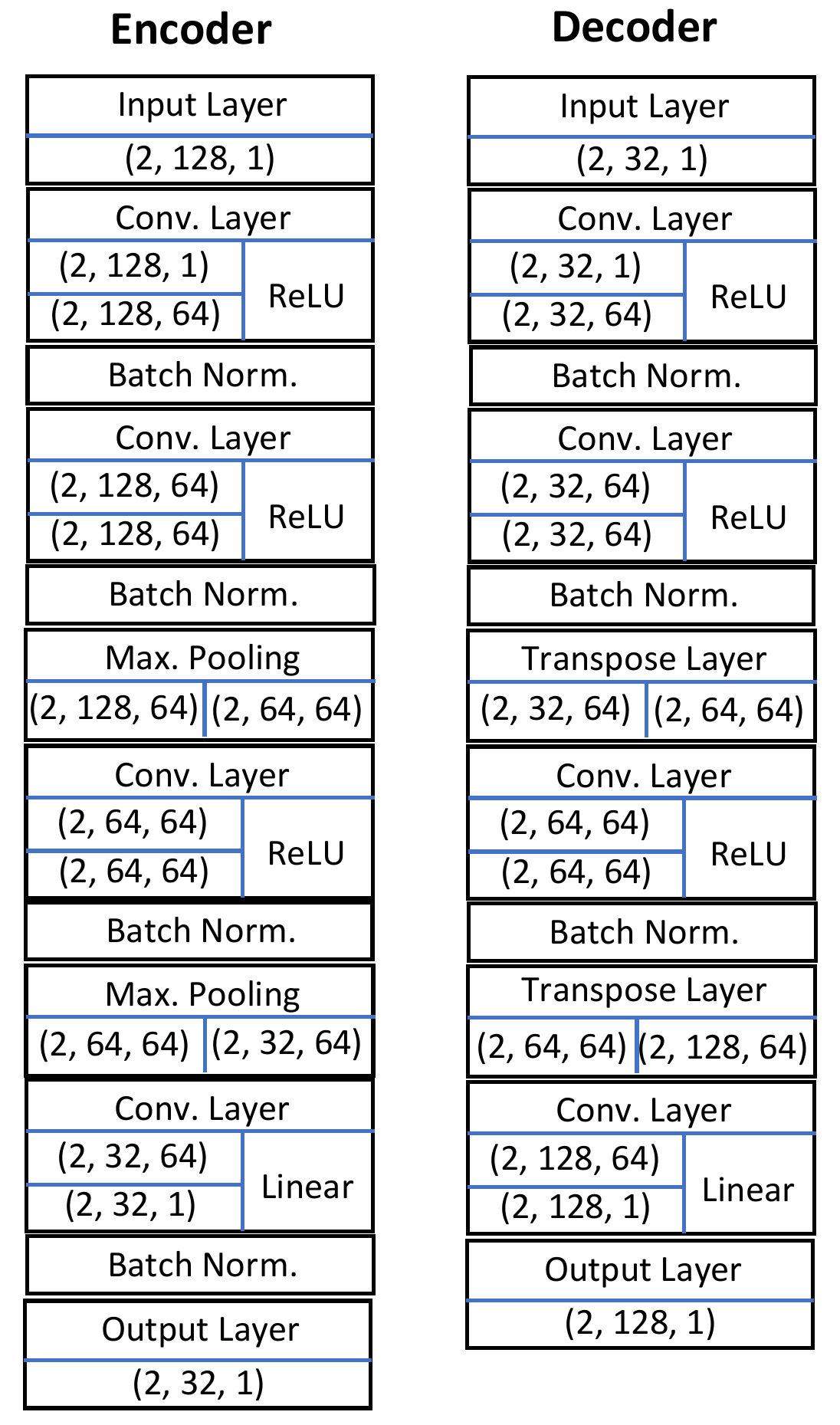}
	\caption{Implementation of the AAE architecture: Encoder and Decoder models in Figure~\ref{fig:apm1}.}\label{fig:apm2}
\end{figure}

\begin{figure}[t!]
	\includegraphics[scale=0.47]{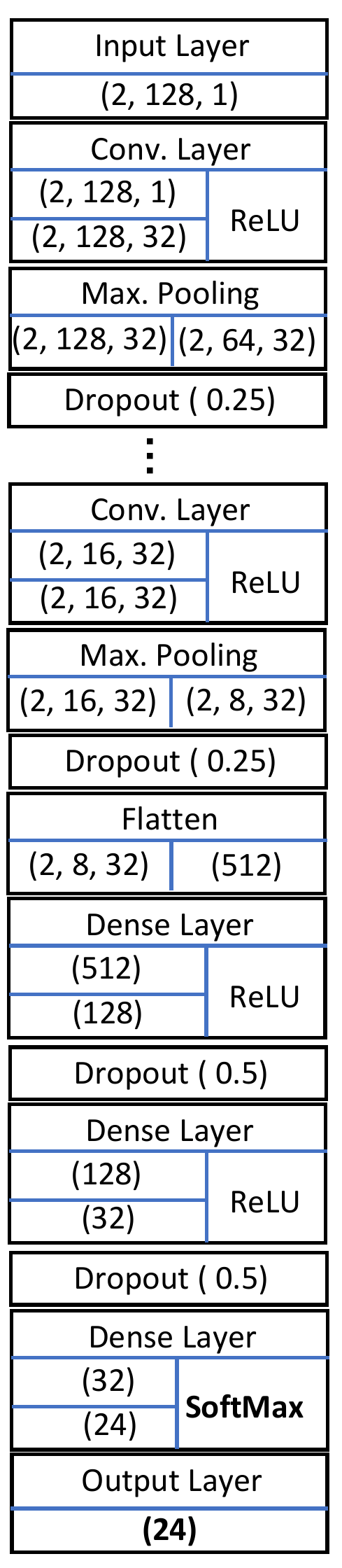}
	\caption{Implementation of the the \textit{DecReg} architecture in Figure~\ref{fig:apm1} (we have the same structure for \textit{EncReg} and \textit{ActReg}).}\label{fig:apm3}
\end{figure}

The Encoder Regularizer (\textit{EncReg}) and the  Activity Regularizer (\textit{ActReg}) share the same architecture as the Decoder Regularizer (\textit{DecReg}). The only differences are that the shape of input for \textit{EncReg} is $32$, instead of $128$, and the shape of softmax output for \textit{ActReg} is $4$, instead of $24$ for a dataset with 24 users and 4  activities. Figure~\ref{fig:apm1} shows the overall architecture whereas Figures~\ref{fig:apm2} and \ref{fig:apm3} show the details of each neural network model. 

Because convolutional layers capture well locally autocorrelated and translation-invariant patterns in time series~\cite{lecun1995convolutional}, we choose the two privacy regularizers, \textit{EncReg} and \textit{DecReg}, and the activity regularizer, \textit{ActReg},  to be convolutional neural network classifiers trained by a categorical cross-entropy loss function~\cite{yang2015deep}.

\subsection{Training}

Instead of just training on a single epoch, as usually done in adversarial training~\cite{goodfellow2014generative}, all the classifiers should be trained for several epochs, $e$, on the entire dataset to converge to suboptimal information estimators for use in the next step. In fact, our objective is not to learn the data distribution, but to transform data from an identity-centred sample space (which is informative about users' identity) to an activity-centred sample space (which carry only information about the underlying activity). Therefore, each regularizer should at least converge to a suboptimal approximator of mutual information.

\begin{figure*}
\begin{algorithmic}[1]
    \algrule
\Procedure{TrainAAE}{$\mathcal{X}, \mathcal{U}, \mathcal{T}$, $e$} \Comment{$\mathcal{X}$: dataset (${M}\times {W}$ temporal windows); $\mathcal{U}$: identity labels; $\mathcal{T}$: activity labels;  $e$ number of epochs.}
    \State \textit{AAE} (\textit{Encoder}+\textit{Decoder}) $\gets$ Random initialization;
    
    \State \textit{AAE} $\gets$ Train on $\mathcal{X}$ as both input and output for $e$ epochs; 
   
    \State $\mathcal{Y} \gets$ \textit{Encoder}($\mathcal{X}$); \Comment{$\mathcal{Y}$ is the extracted latent representation from the raw data.}
    
    \State $\mathcal{X'} \gets CopyOf(\mathcal{X})$; \Comment{Keep raw data intact to use it for evaluation in each iteration.}
    
    \State \textit{EncReg}, \textit{DecReg}, \textit{ActReg}, \textit{AAE} $\gets$ Random initialization;
    
    \Do 
        \State \textit{EncReg} $\gets$ Train on $\mathcal{Y}$ as input and $\mathcal{U}$ as output using categorical cross-entropy as loss function, for  $e$ epochs;
        
        \State \textit{DecReg} $\gets$ Train on $\mathcal{X'}$ as input and $\mathcal{U}$ as output using categorical cross-entropy as loss function, for  $e$ epochs;
        
        \State \textit{ActReg} $\gets$ Train on $\mathcal{X'}$ as input and $\mathcal{T}$ as output using categorical cross-entropy as loss function, for  $e$ epochs;
        
        \State Freeze parameters of \textit{EncReg}, \textit{DecReg}, and \textit{ActReg};
        
        
        \State \textit{AAE} $\gets$  Train on $\mathcal{X'}$ as input and $\mathcal{U}$, $\mathcal{U}$, $\mathcal{T}$, and ${\mathcal{X}}$ as outputs for  $e$ epochs (see Figure \ref{fig:framework});

        \State $\mathcal{Y} \gets$ Encoder(${\mathcal{X}}$);
        \State $\mathcal{X'} \gets$ Decoder($\mathcal{Y}$);
        \State Unfreeze parameters of \textit{EncReg}, \textit{DecReg}, and \textit{ActReg};
    \doWhile{ it does not satisfies the convergence conditions;}
    \State \textbf{return} AAE; \Comment{Resulting AAE to be used as Anonymizer.}
    \algrule
\EndProcedure
\end{algorithmic}
\caption{The adversarial regularization procedure to train the Anonymizer, $\mathcal{A}(\cdot, \theta^{*})$, using Eq.~(\ref{eq:id_act})}
\label{alg:cae}
\end{figure*}

The \textit{EncReg} learns to identify a user among $N$ in the training dataset by getting as input $\mathbf{Y}$ the low-dimensional representation of $\mathbf{X}$ produced by the \textit{Encoder}. The output is the identity label, $\mathbf{U}$. The \textit{DecReg} learns to identify users by getting as input the reconstructed data, $\mathbf{X'}$, produced by the Decoder  (here  too the output is the identity label, $\mathbf{U}$). The  \textit{ActReg} learns to recognize the current activity and gets the reconstructed data, $\mathbf{X'}$, as input and the activity label, $\mathbf{T}$, as output.
Finally, the distortion regularizer, a loss function that constrains the allowed distortion on the data, gets the original data, $\mathbf{X}$, and  reconstructed data, $\mathbf{X'}$, to calculate pointwise the \textit{mean squared error} to quantify the amount of distortion. 

After each iteration, we evaluate the convergence condition of the AAE to decide, based on the current utility-privacy trade-off. We discuss more about possible evaluation methods in Section~\ref{sec:uptoff}.

Figure~\ref{alg:cae} summarizes the training of the AAE, which can be done locally, on the user powerful devices; centrally, by a service provider; or a user can download a public pre-trained model and refined it on their own data~\cite{servia2018privacy}.

\subsection{Multi-objective loss function}\label{sec:loss}

After each round of  training of the regularizers,
we freeze their parameters while training the AAE (line 11 of the training procedure, Figure~\ref{alg:cae}).  
A key contributior to the AAE training is our proposed multi-objective loss function, $\mathrm{L}$, which implements the fitness function $\mathrm{F}\left(\mathrm{A}\left(\mathbf{x}\right)\right)$ of Eq.~(\ref{eq:anon}):
\begin{equation}\label{eq:ws}
        \mathrm{L} = \beta_{i}\mathrm{L}_{i}-\beta_{a}\mathrm{L}_{a}+\beta_{d}\mathrm{L}_{d},
\end{equation}
where the regularization parameters $\beta_{a}$, $\beta_{d}$, and $\beta_{i}$ are non-negative, real-valued weights that determine the utility-privacy trade-off.  
${L}_{a}$ and ${L}_{d}$ are \textit{utility losses} that can be customized based on the app requirements (note that $\mathrm{L}_{d}$ is the only available utility loss if there is no target application), whereas ${L}_{i}$ is an \textit{identity loss} that helps the AAE remove user-specific signals.

The categorical cross-entropy loss function for classification, $\mathrm{L}_{a}$, aims to preserve activity-specific patterns\footnote{We can customize $\mathrm{L}_a$ for the task e.g.~using a binary cross-entropy for fall detection~\cite{majumder2017smart}.}:
\begin{equation}
        \mathrm{L}_{a} =  \mathbf{T} \log(\hat{\mathbf{T}}),
\end{equation}

where $\mathbf{T}$ is the one-hot $B$-dimensional vector of the true activity label for $\mathbf{X}$ and $\hat{\mathbf{T}}$, the output of a softmax function, is a $B$-dimensional vector of probabilities for the prediction of the activity label.

To tune the desired privacy-utility trade-off,  the distance function that controls the amount of distortion, $\mathrm{L}_{d}$, forces $\mathbf{X'}_{{sj}}$ to be as similar as possible to the input $\mathbf{X}_{{sj}}$:
\begin{equation}
        \mathrm{L}_{d} = \frac{1}{M\times W} \sum_{s=1}^{M}\sum_{j=1}^{W} (\mathbf{X}_{{sj}}-\mathbf{X'}_{{sj}})^2,
\end{equation}

Finally, the identity loss, ${L}_{i}$, the most important term of our multi-objective loss function that aims to minimize sensitive information in the data, is defined as:
\begin{equation} \label{eq:lid}
        \mathrm{L}_{i} =
        -\left(
            \mathbf{U} \log
            \left(
                \textbf{1}^{N}- \hat{\mathbf{U}}
            \right)
            +
            \log
            \left(
                1-max
                \left(
                    \hat{\mathbf{U}}
                \right)
            \right)
        \right),
\end{equation}
where $\textbf{1}^{N}$ be the all-one column vector of length $N$, $\mathbf{U}$ is the true identity label for  $\mathbf{X}$, and $\hat{\mathbf{U}}$ is  the output of the softmax function, the $N$-dimensional vector of probabilities  learned by the classifier (i.e.~the probability of each user label, given the input). 

A trivial anonymization would consistently transform data of a user  into the data of another user (and vice versa). However, this transformation would only  satisfy the first element of $L_{i}$. As no attacker should be able to confidently predict $\mathbf{U}$ from $\mathbf{X'}$, we maximize the difference between the prediction, $\hat{\mathbf{U}}$, and the true identity, $\mathbf{U}$ by  minimizing the cross-entropy between the true identity label and the regularizer's prediction of this label, as well as the maximum value of the predicted identity vector, $\mathbf{\hat{U}}$ (see Eq.~(\ref{eq:lid})). The derivation of ${L}_{i}$ is presented in the next section.  

\subsection{Derivation of the identity loss} \label{sec:idloss}


Our goal is to make $\mathbf{U}$ and $\mathbf{X'}$ independent of each other. To this end, we minimize the amount of information leakage from $\mathbf{U}$ to $\mathbf{X'}$~\cite{rassouli2018optimal}. As a function $\mathrm{{\it f}}$ that aims to infer the identity of a user does not increase the available information, the following inequality holds:
\begin{equation}
    \mathrm{I}(\mathbf{U} ; {\mathbf{X'}}) \geq \mathrm{I}(\mathbf{U} ; \mathrm{f}(\mathbf{X'})),
\end{equation}
and therefore if we reduce the mutual information between the user's identity and their released data, the processing of these data cannot increase the mutual information. The mutual information, $\mathrm{I}(\mathbf{U} ; \mathbf{X'})$, can be defined as
\begin{equation}
\mathrm{I}(\mathbf{U} ; \mathbf{X'}) = \mathrm{H}(\mathbf{U})- \mathrm{H}(\mathbf{U} \vert  \mathbf{X'}),
\end{equation}
where $\mathrm{H}(\cdot)$ is the entropy. As the entropy is 
non-negative and we cannot control  $\mathrm{H}(\mathbf{U})$, we maximize  the conditional entropy between identity variable and the transformed data, $\mathrm{H}(\mathbf{U} \vert \mathbf{X'})$, in order to minimize the mutual information, $\mathrm{I}(\mathbf{U} ; \mathbf{X'})$:
\begin{equation}
\mathrm{H}(\mathbf{U} \vert \mathbf{X'}) = \mathrm{H}(\mathbf{U} , \mathbf{X'}) - \mathrm{H}(\mathbf{X'}).
\end{equation}

The entropy of $\mathbf{X'}$, $\mathrm{H}(\mathbf{X'})$, can be reduced independently of any other latent variables by simply downsampling the data. However, as blindly minimizing $\mathrm{H}(\mathbf{X'})$ could lead to a substantial utility loss, we focus on maximizing $\mathrm{H}(\mathbf{U} , \mathbf{X'})$.

Let $p(\mathbf{U},\mathbf{X'})$ be the joint distribution of $\mathbf{U}$ and    $\mathbf{X'}$; and $S_\mathbf{u}$ and $S_\mathbf{X'}$ be the supports of $\mathbf{U}$ and $\mathbf{X'}$, respectively. Then%
\begin{equation}
    \mathrm{H}(\mathbf{U} , \mathbf{X'} )= -\int_{S_\mathbf{u}}\int_{S_\mathbf{\mathbf{X'}}} p(\mathbf{U},\mathbf{X'})\log{p(\mathbf{U},\mathbf{X'})}.
\end{equation}

We now need an estimator for $\mathrm{H}(\mathbf{U}, \mathbf{X'})$ as we cannot calculate the joint entropy directly for high-dimensional data. When labeled data are available,  $\mathbf{X'}$ can be used as input to predict $\hat{\mathbf{U}}$ as an estimation of $\mathbf{U}$. We therefore reformulate the problem of maximizing the joint entropy, $\mathrm{H}(\mathbf{U}, {\mathbf{X'}})$, as  maximization of the cross entropy between the true label, $\mathbf{U}$, and the predicted label, $\hat{\mathbf{U}}$:
\begin{equation}
    \mathrm{H}_{{\hat{\mathbf{U}}}}({\mathbf{U}}) = - \int_{S_{{\mathbf{\mathbf{X'}}}}} {\mathbf{U}}\log{\hat{\mathbf{U}}}.   
\end{equation}

If $\hat{\mathbf{U}}[{k}]$ is the ${k}$-th element of the vector predicted by the multiclass classifier, the empirical cross entropy for data ${\mathbf{X'}}$ of user ${k}$ is:
\begin{equation}
    - \mathbf{U}\log{\hat{\mathbf{U}}} = - \log{\hat{\mathbf{U}}[{k}]} 
\end{equation}
and, since $\hat{\mathbf{U}}[{k}] \in [0,1]$, maximizing $ - \log{\hat{\mathbf{U}}[{k}]}$ is equivalent to minimizing $ - \log{({1}-\hat{\mathbf{U}}[{k}])}$. Therefore minimizing the first term of Eq.~(\ref{eq:lid}), $\mathbf{U} \log(\textbf{1}^{N} - \hat{\mathbf{U}})$, minimizes the mutual information, $\mathrm{I}(\mathbf{U};\mathbf{X'})$, and, by forcing the AAE to minimize this value, we minimize the amount of user-identifiable information in $\mathbf{X}'$.

\subsection{Examples}\label{sec:exmpl}

To gain an appreciation of the type of distortions introduced by the AAE, we compare sensor data before and after transformation.

\begin{figure}[t!]
	\begin{minipage}[t]{.92\linewidth}
		\includegraphics[width=\linewidth]{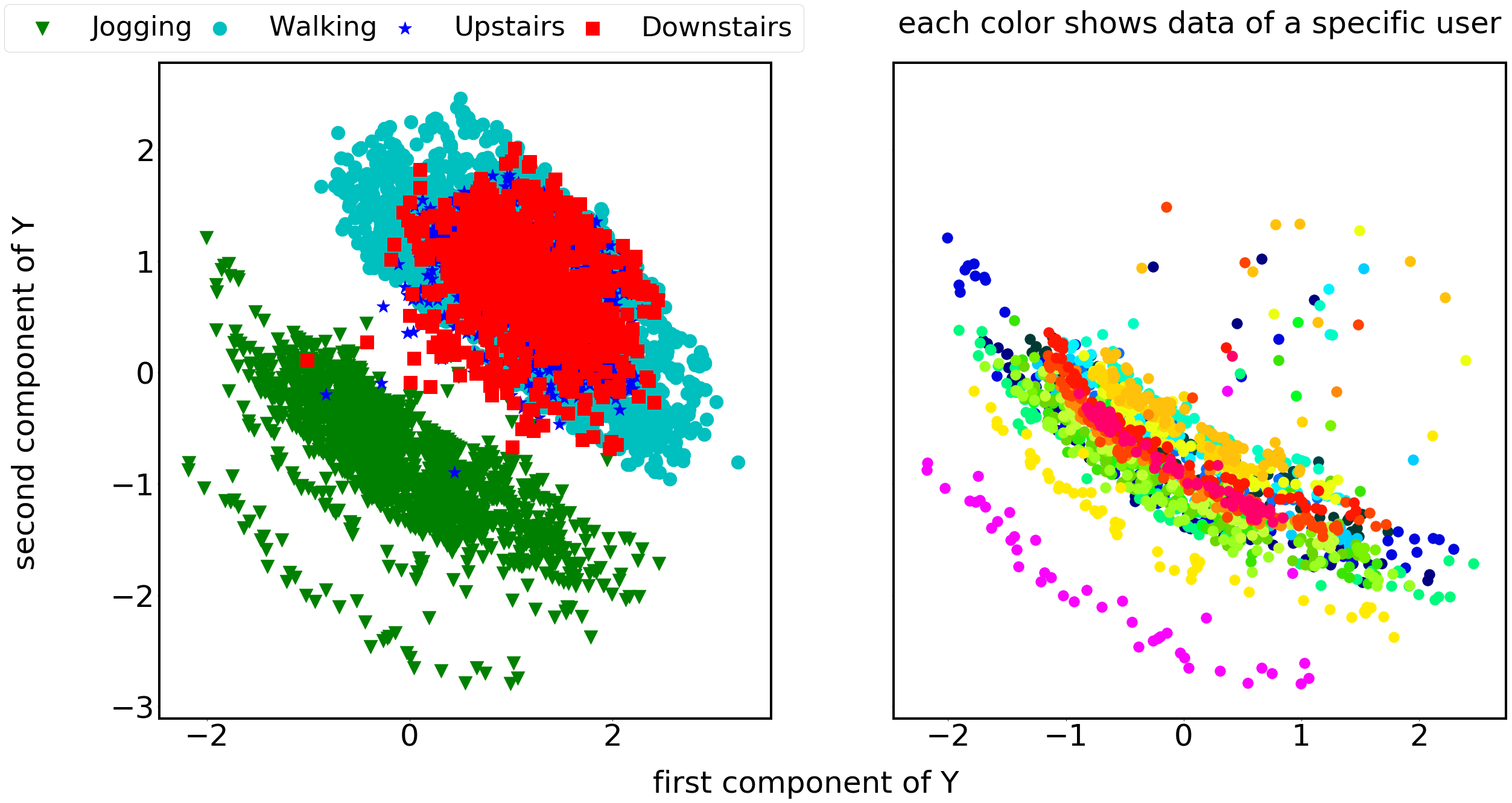}
	\end{minipage}
	\vfill
	\begin{minipage}[t]{.92\linewidth}
		\includegraphics[width=\linewidth]{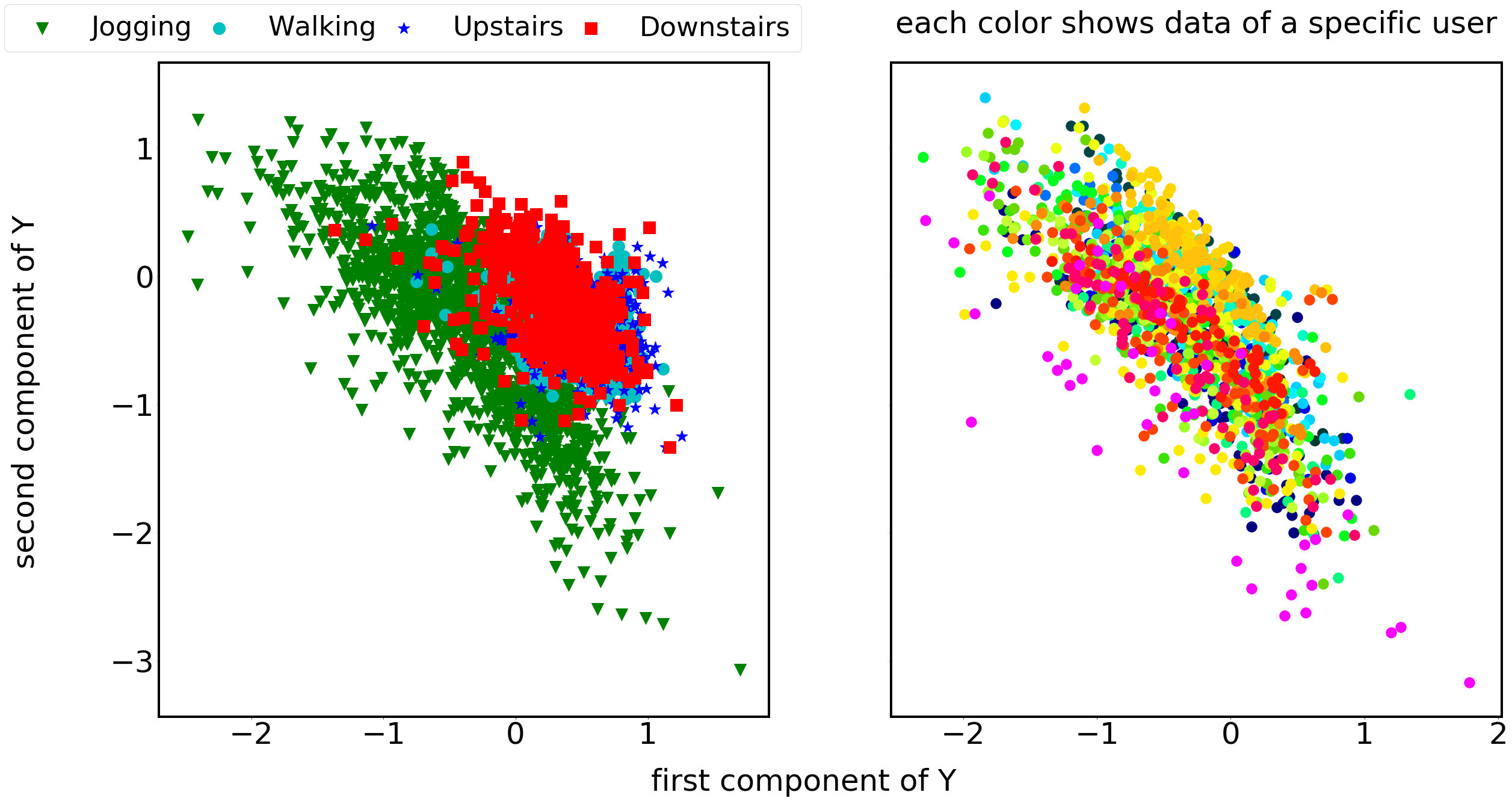}
	\end{minipage} 
 \caption{Latent representation, $\mathbf{Y}$, of the 64D  gyroscope data in 2D. (Top row): raw data. (Bottom row): data transformed by the AAE. \textbf{(Left column)}: samples of four  activities. \textbf{(Right column)}: Jogging data for all users. }
    \label{fig:y2comps}
\end{figure}

Figure~\ref{fig:y2comps} (top) shows the low-dimensional latent representation of raw gyroscope data extracted by the bottleneck of the model. The distribution of $\mathbf{Y}$ has useful information to distinguish not only the activities, but also the users (color clusters of the top-right plot). 
Figure~\ref{fig:y2comps} (bottom) shows the latent representation of the data anonymized  by our method: the  transformation masks the data  for different users but preserves the Jogging activity samples separated from those of the other activities (note that this is a considerably compressed representation of the input data). 

\begin{figure}[t]
	\centerline{\includegraphics[scale=0.07]{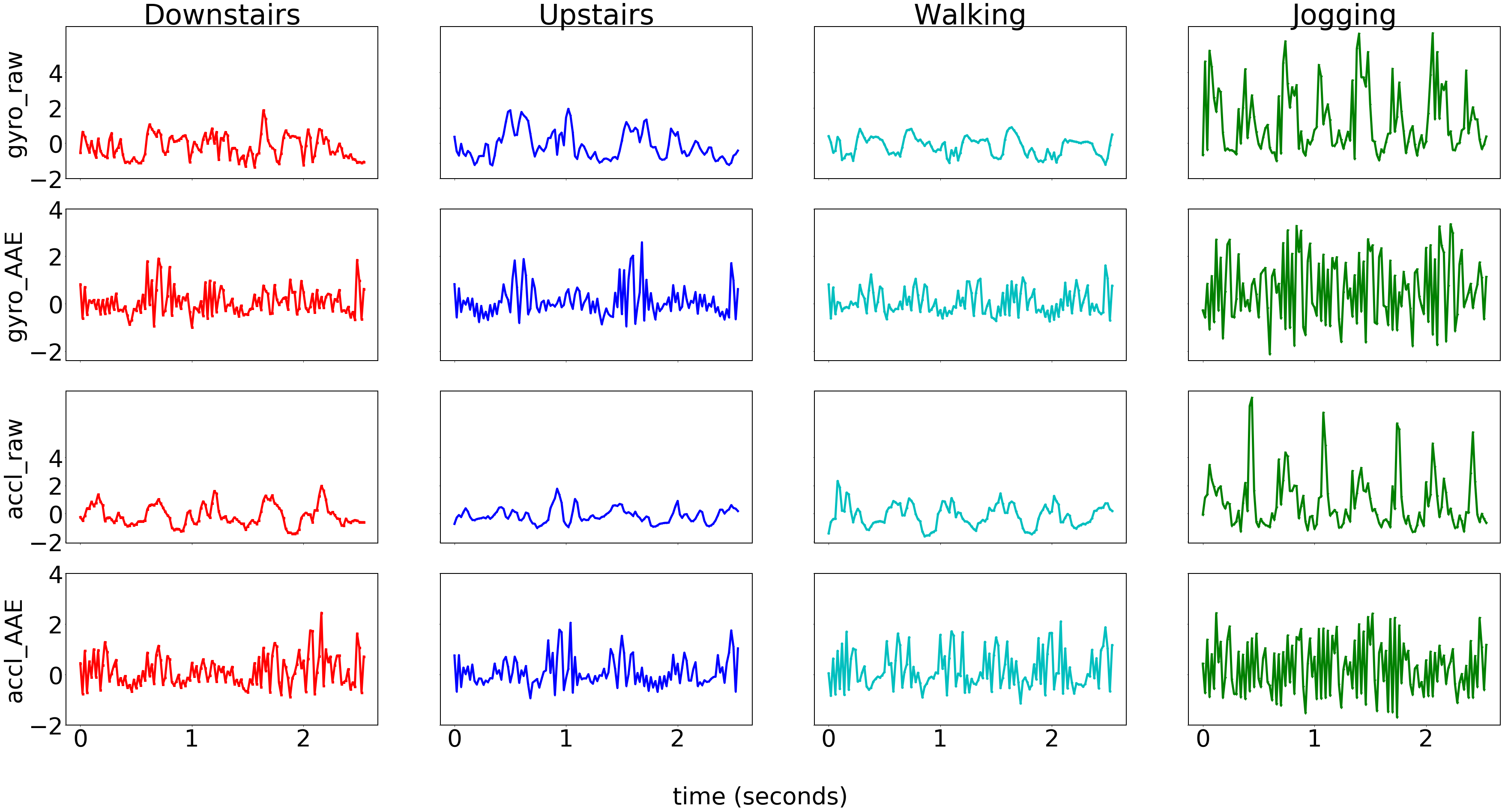}}
	\caption{Comparison of raw  (first and third row) and transformed data (second and fourth row) for gyroscope (first two rows) and accelerometer (last two rows) for four activities.}
	\label{fig:raw_cae}
\end{figure}

Figure~\ref{fig:raw_cae} compares raw and transformed data of four activities. It is possible to notice that the AAE obscures patterns and peaks, but maintains differences among data of different activities. 

\begin{figure}[t]
	\centerline{\includegraphics[scale=0.067]{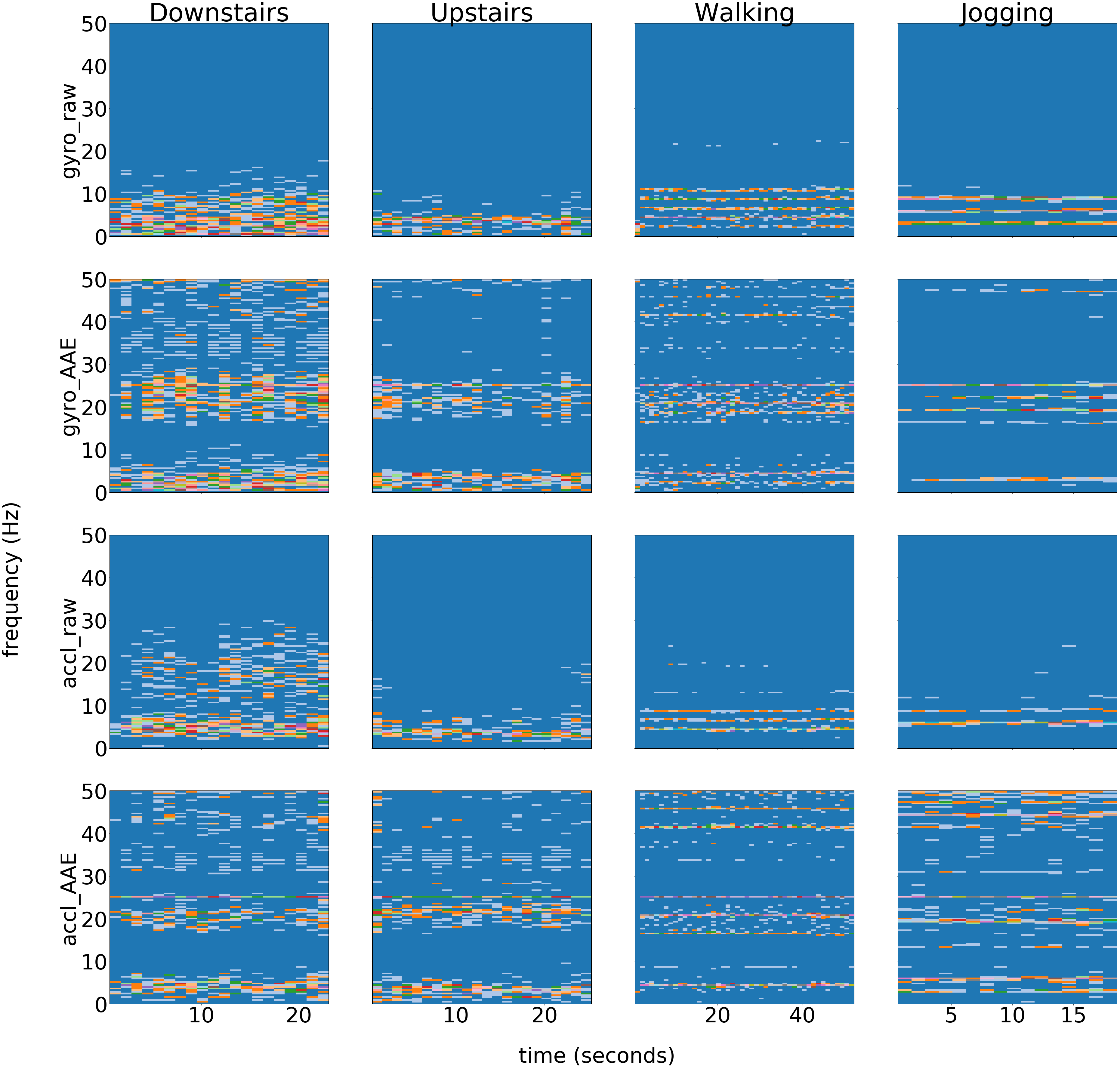}}
	\caption{Spectrogram of raw  (first and third row) and transformed data (second and fourth row) for gyroscope (first two rows) and accelerometer (last two rows) for four activities.}
	\label{fig:freq}
\end{figure}

Finally, Figure~\ref{fig:freq} compares the spectrogram of raw and transformed data for a user: the AAE introduces new periodic components and obscure some of the original ones, and they differ across the activities. As periodic components in accelerometer data can disclose information about  attributes  of users such as  height and weight, the AAE reduces the possibility of user re-identification by introducing new periodic components in the data.

In the next section we quantify the performance of the proposed method and compare it with alternative approaches.

\section{Evaluation} 
\label{sec:eval}

To evaluate the effectiveness of the proposed data anonymizer, we analyze the trade-off between recognizing the activity of a user  and concealing their identity.  We measure the extent to which the activity recognition accuracy is reduced by the anonymization process, compared to using the raw data. We compare with two baseline methods for coarse-grained time series data, namely  \textit{Resampling} and \textit{Singular Spectrum Analysis} (SSA), and with \textit{REP}~\cite{edwards2015censoring}, which only considers sensitive information included in $\mathbf{Y}$ and does not take $\mathbf{X}'$ into account~(Figure~\ref{fig:framework}). 




\subsection{Experimental Setup} \label{sec:ex_set}

Current public datasets of motion sensor data do not simultaneously satisfy the requirements of abundance and variety of activities and users\footnote{Datasets that satisfy both (e.g.~\cite{neverova2016learning}) are still private.}. We therefore collected a dataset from the accelerometer and gyroscope of an iPhone 6s placed in the user's front pocket of tight trousers~\cite{malekzadeh2018protecting, katevas2014poster}. The dataset includes 24 participants, in a range of age, weight, height and gender, who performed 6 activities in 15 trials. In each trial, we used the same environment and conditions for all the users~(see Table~\ref{tab:dataset}). 
\begin{table}
	{
	
	\begin{tabular}{l|l}
\hline
Number of users & 24 (14 males, 10 females) \\ \hline
Sampling rate & 50 Hz \\ \hline
Sensors & \begin{tabular}[c]{@{}l@{}}gyroscope\\ accelerometer\end{tabular} \\ \hline
Features & \begin{tabular}[c]{@{}l@{}}rotationRate (x,y,z)\\ userAcceleration (x,y,z)\\ gravity (x,y,z)\\ attitude(roll, pitch, yaw)\end{tabular} \\ \hline
\begin{tabular}[l]{@{}l@{}}Activities\\ (number of trials)\end{tabular} & \begin{tabular}[c]{@{}l@{}}Downstairs (3 trials )\\ Upstairs (3 trials)\\ Walking (3 trials)\\ Jogging (2 trials)\\ Sat (2 trials)\\ Stand-Up (2 trials)\end{tabular} \\ \hline
\end{tabular}
	
   \caption{{
		 The MotionSense dataset~\cite{malekzadeh2018protecting}. Multiple trials of the same activity are performed in  different locations. KEY -- (x, y, z): the three  axes of the sensor.  }
	}
	\label{tab:dataset}
}
\end{table}
We divide the dataset into training and test sets with two different strategies, namely {Subject} and {Trial}. In \textit{Subject}, we use as test data all the data of $4$ users, $2$ females and $2$ males, and as training data that of the remaining $20$ users. After training,  the model is evaluated on data of $20$ unseen users. In \textit{Trial}, we use as test data one trial session for each user and as training data the remaining trial sessions (for example, one trial of Walking of each user is used as test and the other two trials are used as training). In both cases, we put $20\%$ of training data for validation during the training phase. We repeat each experiment 5 times and report the mean and the standard deviation. For all the experiments we use the magnitude value for both gyroscope and accelerometer.

We choose as window length ${W}=128$ (2.56 seconds) and we set as stride ${S}=10$. 
For all the regularizers, \textit{EncReg}, \textit{DecReg}, and \textit{ActReg}, we use 2D convolutional neural networks. To prevent overfitting to the training data, we put a Dropout~\cite{srivastava2014dropout} layer after each convolution layer. We also use an L2 regularization to penalize large weights so that the classifier is forced to learn features that are more relevant for the prediction.

\subsection{Sensor Data Characteristics}
\label{sec:motiv}

In this section we discuss the characteristics of motion sensor data that informed the design of our sensor data anonymizer.

\begin{figure}[t]
\centerline{\includegraphics[scale=0.2]{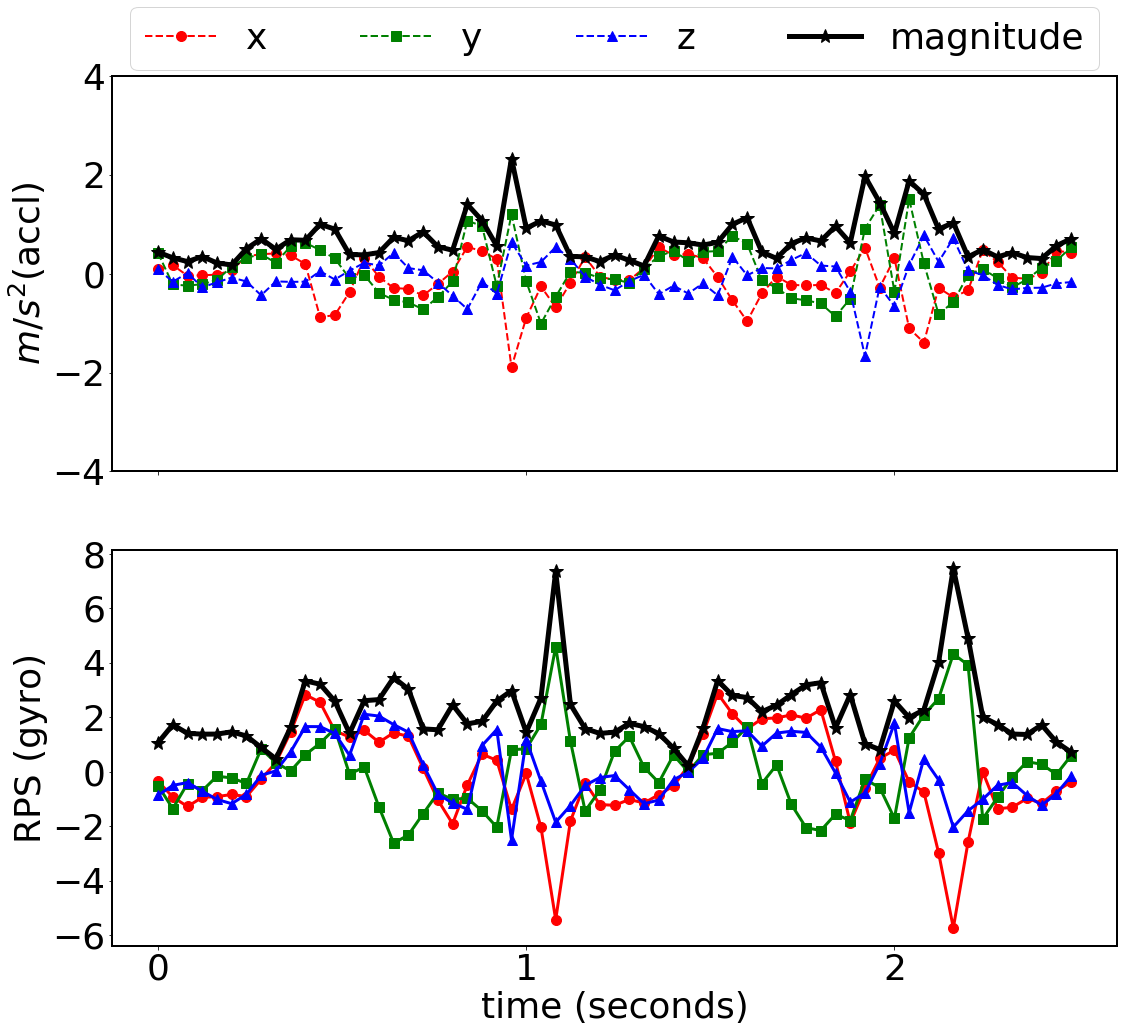}}
 \caption{Sample accelerometer~(top) and gyroscope~(bottom) data for Walking of a specific user. KEY -- RPS: revolutions per second, $m/s^2$: metres per second squared}
    \label{fig:raw_mag}
\end{figure}

\begin{figure}[t]
\centerline{\includegraphics[scale=0.17]{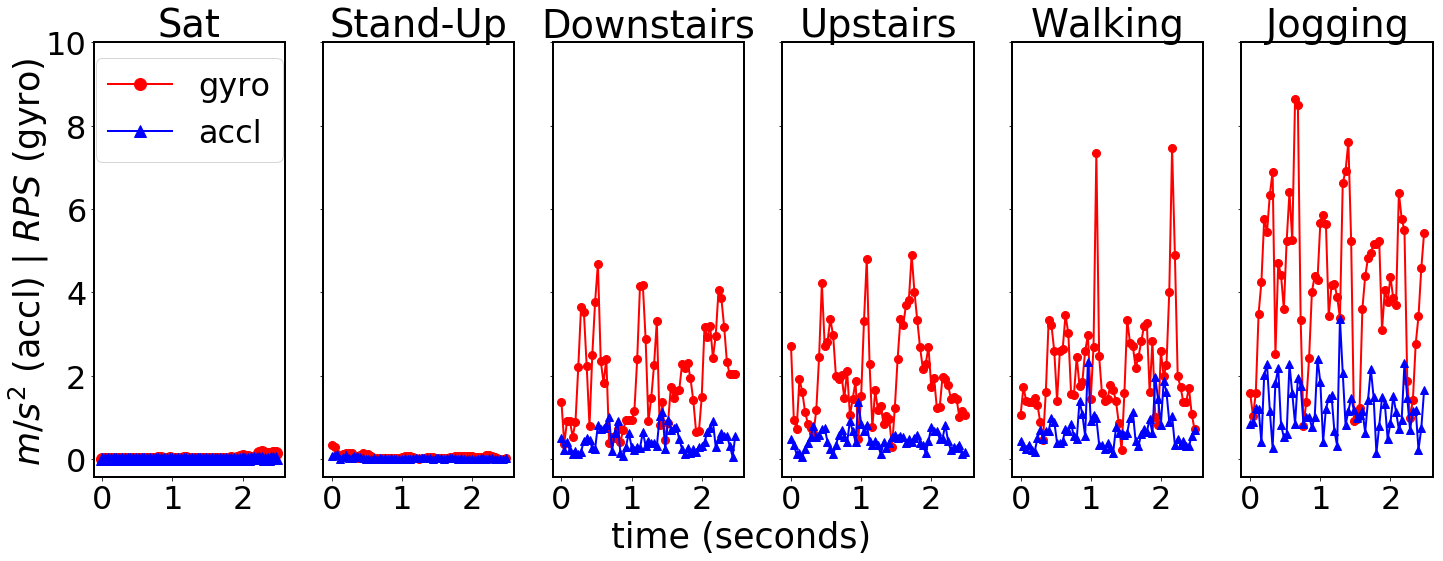}}
 \caption{Sample  accelerometer~(accl) and gyroscope~(gyro) data for six activities for a single user.}
    \label{fig:six_act}
\end{figure}

 Figure~\ref{fig:raw_mag} shows the correlation between the magnitude of the time series collected from these sensors. We see that both sensors almost follow each other, especially for the peaks and periodicity of the magnitude value, whereas a correlation among axes is less obvious.
Figure~\ref{fig:six_act} compares the magnitude values of the data from two sensors when the user performs in six different activities. Note that Sat and Stand-Up are difficult to be told apart. The only data that are informative to distinguish these activities from each other are the values of the gravity axes which determine whether the phone is held  vertically or horizontally. However, we do not consider Sat and Stand-Up in our experiments for training the AAE.

\begin{figure}[t]
	\begin{minipage}[t]{.99\linewidth}
		\includegraphics[width=\linewidth]{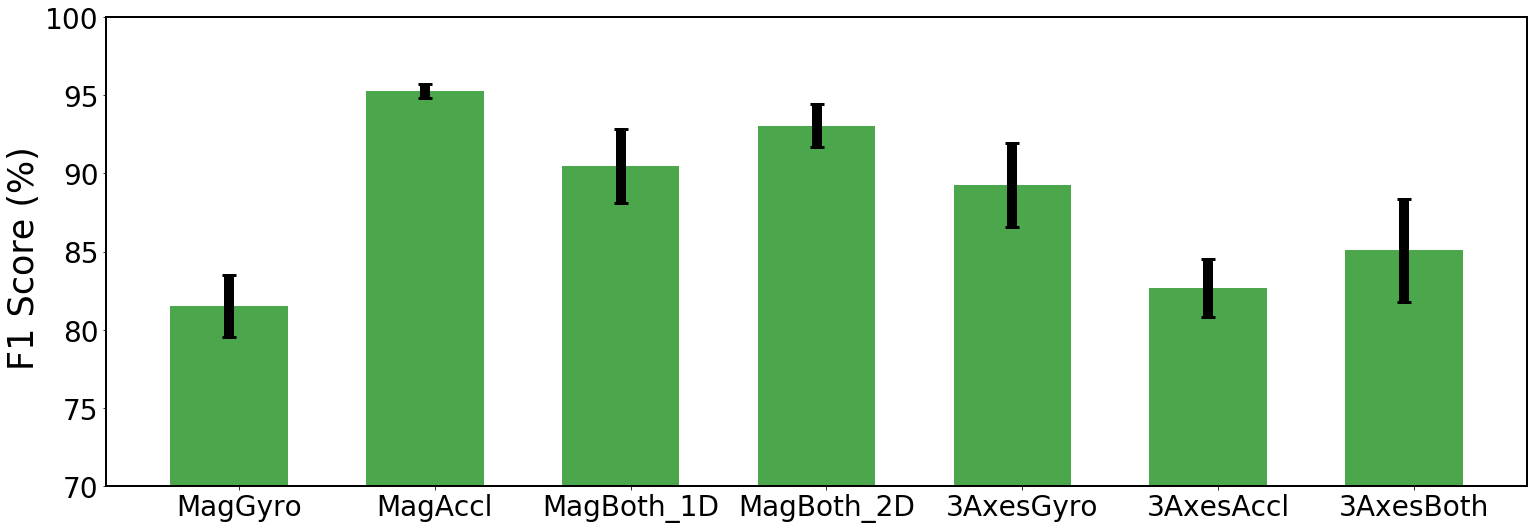}
	\end{minipage}
	\vfill
	\begin{minipage}[t]{.99\linewidth}
		\includegraphics[width=\linewidth]{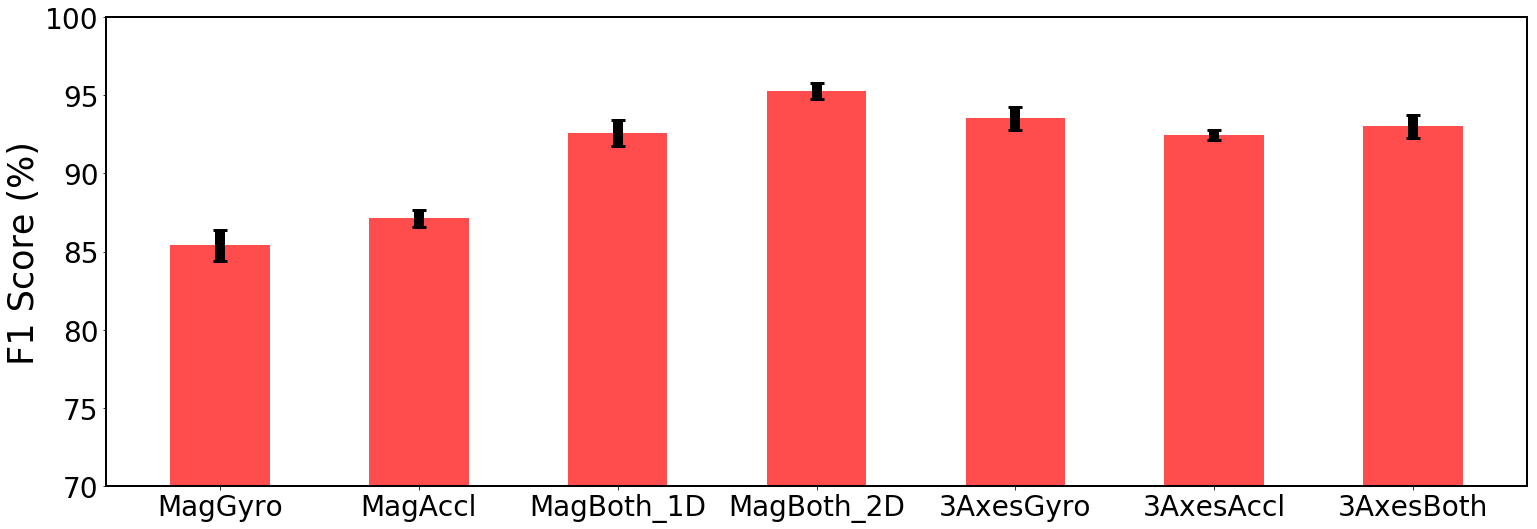}
	\end{minipage} 
 \caption{Average F1 score for the recognition, with different sensor data types, of activity (top) and identity (bottom).  The black vertical segments show the standard deviation. KEY -- {Mag}: magnitude; gyro: gyroscope; accl: accelerometer; {Both}: both  gyro and accl; {1D} and {2D} are the dimensions of the convolution filter.}
    \label{fig:sens_dt}
\end{figure}

Figure~\ref{fig:sens_dt} compares the F1 score obtained using as classifier a deep convolutional neural network with seven groups of data\footnote{By the similar architecture described in Figure~\ref{fig:apm3}}. We use the {\em Subject} setting for activity recognition and the {\em Trial} setting for identity recognition. The groups of data are the magnitude value of each sensor, the exact value of each axis, the data of only one of these sensors and then both. It is interesting to estimate the amount of information about user's identity that can be extracted form the correlation between accelerometer and gyroscope. Note that we can achieve equal (or better) accuracy for activity recognition using only the magnitude, whereas we should use the values of each axis for identity recognition. Moreover, using a 2D convolutional filter (i.e.~the classifier considers the correlation among the input sensors) improves over both activity and identity recognition compared to using 1D filters, which process each input separately. Hence, a good anonymization mechanism should consider both inter-sensor and  intra-sensor correlations. We will use the magnitude value of both gyro and accelerometer (${MagBoth\_2D}$) in our experiments of evaluating the utility-privacy trade-offs.

\begin{figure}[t]
\centerline{\includegraphics[scale=0.22]{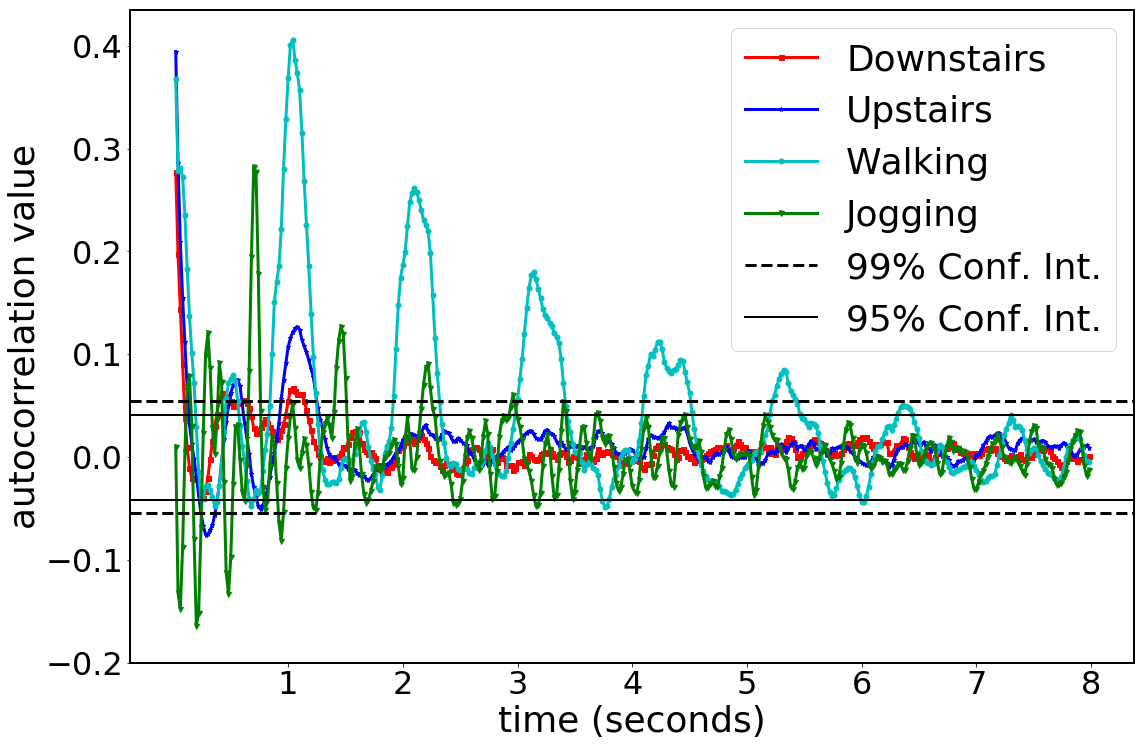}}
 \caption{Autocorrelation  of accelerometer data for four activities averaged over all the users. Correlation values outside the lines of the confidence interval  (Conf.~Int.) are statistically significant.}
    \label{fig:acf}
\end{figure}

Figure~\ref{fig:acf} shows the autocorrelation at varying time lags for the magnitude of accelerometer data for different activities (average over 45 seconds of data for all users). Note that each activity has a different period. Walking has the highest correlation, followed by Jogging, Upstairs, and Downstairs. The distance between two peaks can be related to the stride. There are also strong correlations among samples inside a 2-second window, whereas correlations go under the confidence interval after about 5 seconds.

\begin{figure}[t]
\centerline{\includegraphics[scale=0.22]{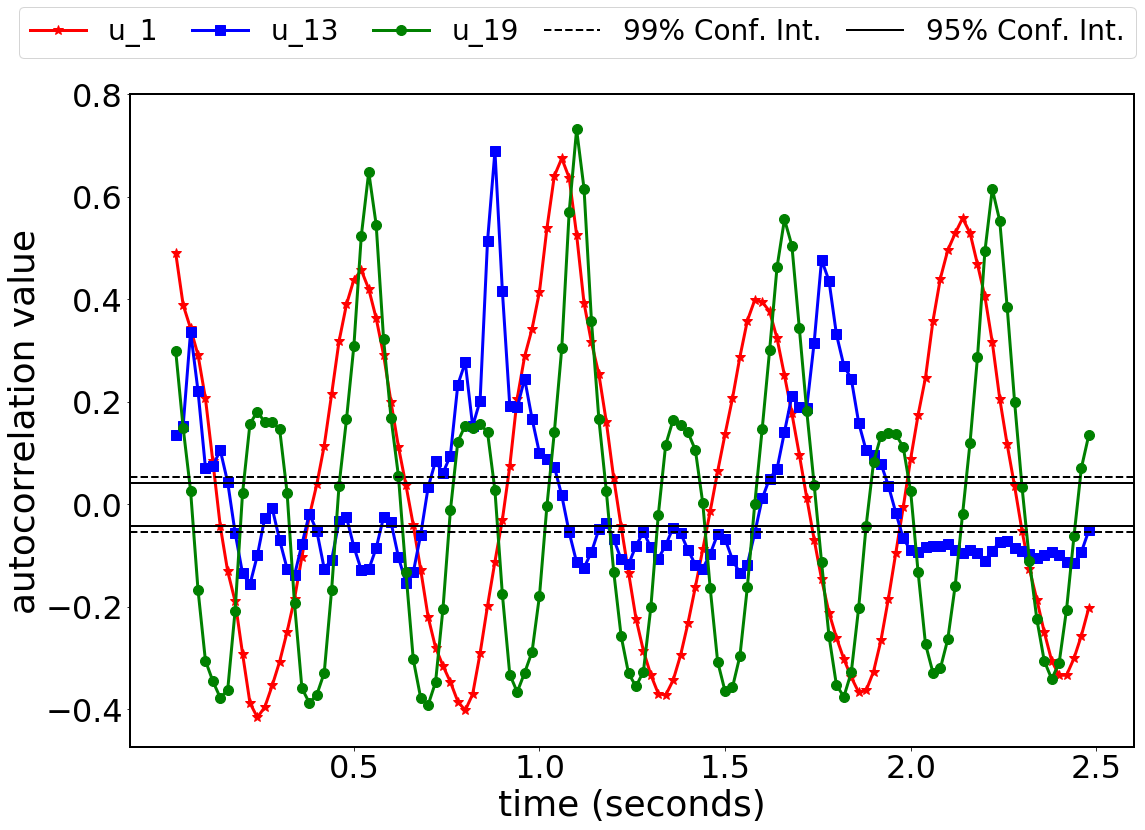}}
 \caption{Autocorrelation of the  accelerometer data for Walking for three users. KEY -- Conf.Int.: confidence interval; $u_1$, $u_{13}$, $u_{19}$: data of user 1, 13, and 19.}
    \label{fig:acf_3sub}
\end{figure}

Figure~\ref{fig:acf_3sub} shows the autocorrelations of the same activity performed by three users. The heavier the user, the longer the intervals between two peaks (user $\mathbf{u}\_1$ is the heaviest among the three). This user-identifiable pattern is a challenging feature to obscure before sharing the data. In fact, we see that baseline methods like downsampling cannot hide the user identity.
%

\begin{table*}[t]
\small
	\begin{tabular}{|l|l|c|c|c|c|c|c|c|}
\hline
 info. & result & \begin{tabular}[c]{@{}c@{}}raw\\  (50Hz)\end{tabular} & \begin{tabular}[c]{@{}c@{}}resample \\ (10Hz)\end{tabular} & \begin{tabular}[c]{@{}c@{}}resample \\ (5Hz)\end{tabular} & \begin{tabular}[c]{@{}c@{}}SSA\\  (1,2)\end{tabular} & \begin{tabular}[c]{@{}c@{}}SSA \\ (1)\end{tabular} & \begin{tabular}[c]{@{}c@{}}REP~\cite{edwards2015censoring} \\ (50Hz)\end{tabular} &
 \begin{tabular}[c]{@{}c@{}}\textbf{AAE} \\ (50Hz)\end{tabular} \\ \hline
\multirow{2}{*}{\begin{tabular}[l]{@{}l@{}} ACT\end{tabular}} & mean F1 & 92.51 & 91.11 & 88.02 & 88.59 & 87.41 & 91.47 &  \textbf{92.91} \\ \cdashline{2-9} 
 & var F1 & 2.06 & 0.63 & 1.85 & 0.91 & 0.89 & 00.87 & \textbf{0.37} \\ \hline
\multirow{2}{*}{\begin{tabular}[l]{@{}l@{}} ID\end{tabular}} & mean ACC & 96.20 & 31.08 & 13.53 & 34.13 & 16.07 & 15.92 & \textbf{6.98}\\ \cdashline{2-9} 
 & mean F1 & 95.90 & 25.57 & 8.86 & 28.59 & 12.58  & 11.25 & \textbf{1.76} \\ \hline
\multirow{2}{*}{\begin{tabular}[l]{@{}l@{}} DTW\end{tabular}} & mean Rank & 0 & 7.2 & 9.3 & 6.8 & 9.5 & 10.7 & \textbf{6.6} \\ \cdashline{2-9} 
 & var Rank & 0 & 5.7 & 5.8 & 5.6 & 5.4 & 5.5 & \textbf{4.7} \\ \hline
\end{tabular}
\caption{Trade-off between utility (activity recognition) and privacy (identity recognition). KEY -- ACT: activity recognition, ID: identity recognition, ACC: accuracy, F1: F1 score, DTW: Dynamic Time Warping as the similarity measure, SSA: Singular Spectrum Analysis,  REP: Only Anonymizing the latent Representation, AAE: Our Anonymizing AutoEncoder. The forth row shows the K-NN rank between 24 users.} 
\label{tab:results}
\end{table*}

\subsection{Baseline Methods}

As baseline methods we use  Resampling and Singular Spectrum Analysis.

Resampling ideally aims to reduce the richness of the data to the extent that it contains useful information for recognizing the  activity but not identity-specific patterns. We choose a resampling based on the {\em Fast Fourier Transform} (FFT) and, specifically, we use the ``signal.resample'' function of ``SciPy'' package~\cite{jones2014scipy}. 
\begin{figure}[t!]
	\begin{minipage}[t]{.495\linewidth}
		\includegraphics[width=\linewidth]{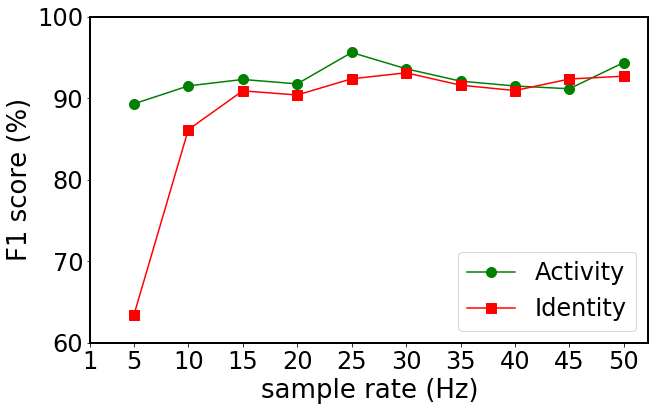}
		\label{fig:resmpl}
	\end{minipage}%
	\hfill%
	\begin{minipage}[t]{.495\linewidth}
		\includegraphics[width=\linewidth]{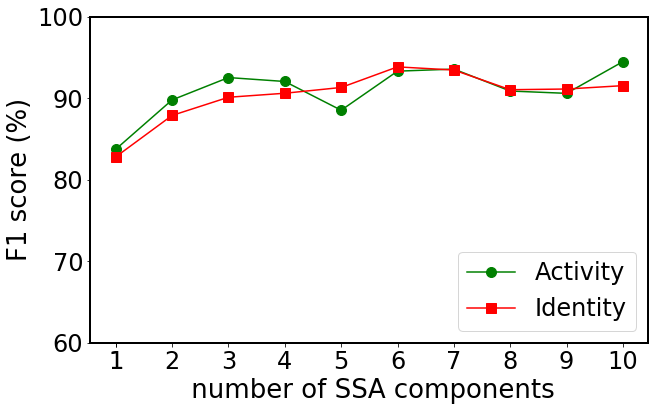}
		\label{fig:ssa}
	\end{minipage} 
	\caption{Classification accuracy for a deep convolutional neural network for both Activity and Identity recognition. \textbf{(Left)}~Using data resampled to another rate (from 5 to 50 Hz, where 50 Hz is the original sampling rate). \textbf{(Right)}~Using data reconstructed using only a subset of components (from 1 to 10, from a total of 50), ordered from largest to smallest by corresponding singular values.
	\label{fig:rs_ssa}}
\end{figure}
Figure~\ref{fig:rs_ssa} (left plot) shows the classification accuracy with downsampled sensor data. For a fair comparison, we trained a fixed model (in terms of the size of the parameters and number of the layers) for all the sample rates. The impact of downsampling on activity recognition can be ignored for a rates greater than 20Hz. However, even at 5Hz, we can distinguish the 24 users from each other with over 60\% accuracy. 

{\em Singular Spectrum Analysis} (SSA)~\cite{ broomhead1986extracting} decomposes time series into interpretable components such as trend, period, and structureless (or noise) components. The window length parameter specifies the number of components. We  decompose each $\mathbf{X}$, into a set of $D$ components, $\{\mathsf{X}_1, \mathsf{X}_2, \ldots, \mathsf{X}_D\}$, such that the original time series can be recovered as:
\begin{equation}
\mathbf{X} = \sum_{d=1}^{D} \mathsf{X}_d.
\end{equation}

As SSA arranges the elements $\mathsf{X}_d$ in descending order according to their corresponding singular value, we explore the idea of {\it incremental reconstruction}. Figure~\ref{fig:rs_ssa} (right plot)  shows that training a classifier on the reconstruction with only the first components, up to the total of 10 extracted components, can achieve over 80\% accuracy for both activity and identity recognition. 

\subsection{Discussion}
\label{sec:uptoff}

In this section, we compare the transformed data produced by our trained AAE with the outputs of the other methods.   

We train an activity recognition classifier on both the raw data and the transformed data, and then use it for inference on the corresponding test data. Here we use the Subject setting, thus the test data includes data of new unseen users. The second row of Table~\ref{tab:results} shows that the average accuracy for activity recognition for both Raw and AAE data is around 92\%. Compared to other methods that decrease the utility of the data, we can preserve the utility and even slightly improve it, on average, as the AAE shapes data such that an activity recognition classifier can learn better from the transformed data than from the raw data. 

To evaluate the degree of anonymity, we assume that an adversary has access to the training dataset and we measure the ability of a pre-trained deep classifier on users raw data in inferring the identity of the users when it receives the transformed data. We train a classifier in the Trial setting over raw data and then feed it different types of transformed data. The third row of
Table~\ref{tab:results} shows that downsampling data from 50Hz to 5Hz reveals more information than using the AAE output in the original frequency. These results show that the AAE can effectively obscure user-identifiable information so that even a model that have had access to users' original data cannot distinguish them after applying the transformation.
 
Finally, to evaluate the efficiency of the anonymization with another unsupervised mechanism, we implement the $k$-Nearest Neighbors ($k$-NN) with Dynamic Time Warping (DTW)~\cite{salvador2007toward}.
Using DTW, we measure the similarity between the transformed data of a target user ${k}$ and the raw data of each user ${l}$, $\mathbf{X}^{l}$, for all ${l} \in \{{1},{2},\ldots,{k},\ldots,{N}\}$. Then we use this similarity measure to find the nearest neighbors of user ${l}$ and check the rank of ${k}$ among them.
The last row of Table~\ref{tab:results} shows that it is very difficult to find similarities between the transformed  and raw data of the users as the performance of the AAE is very similar to the baseline methods and the constraint in Eq.~(\ref{eq:id_act}) maintain the data as similar as possible to the original data.

\section{Conclusion}

We proposed a multi-objective loss function to train an anonymizing autoencoder~(AAE) as sensor data anonymizer for personal and wearable devices. To remove user-identifiable features included in the  data we consider not only the feature extractor of the neural network model~(encoder), but we also force the reconstructor~(decoder) to shape the final output independently of each user in the training set, so the final trained model is a generalized model that can be used by a new unseen user. We ensure that the transformed data is minimally perturbed so an app  can still produce accurate results, for example for activity recognition. The proposed solution is important to ensure anonymization for participatory sensing~\cite{Burke06participatorysensing}, when individuals contribute data recorded by their personal devices for health and well-being data analysis.   

As future work, we aim to measure the cost of running such local transformations on user devices; to conduct experiments on other use cases (i.e.~different tasks); and to derive statistical bounds for the level of privacy protection achieved.

\section*{acknowledgements}
This work was supported by the Life Sciences Initiative at Queen Mary University London and a Microsoft Azure for Research Award (CRM:0740917). Hamed Haddadi was partially supported by the EPSRC Databox grant (EP/N028260/1). Andrea Cavallaro wishes to thank the Alan Turing Institute (EP/N510129/1), which is funded by the EPSRC, for its support through the project PRIMULA. We would also like to thank Deniz Gunduz and Emiliano De Cristofaro for their constructive feedback and insights.

\bibliographystyle{ACM-Reference-Format}

\end{document}